\documentclass[lettersize,journal]{IEEEtran}
\IEEEoverridecommandlockouts
\usepackage{cite}
\usepackage{amsmath,amssymb,amsfonts}
\usepackage{algorithmic}
\usepackage{graphicx}
\usepackage{textcomp}
\usepackage{xcolor}
\usepackage{booktabs}
\usepackage{array}
\usepackage{multirow}
\usepackage{tabularx}
\usepackage{tcolorbox}
\usepackage{makecell}
\usepackage{algorithm}
\usepackage{algorithmic}
\def\BibTeX{{\rm B\kern-.05em{\sc i\kern-.025em b}\kern-.08em
    T\kern-.1667em\lower.7ex\hbox{E}\kern-.125emX}}
\begin{document}

\title{MAction-SocialNav: Multi-Action Socially Compliant Navigation via Reasoning-enhanced Prompt Tuning}

\author{
\IEEEauthorblockN{
    Zishuo~Wang$^{1}$,
    Xinyu~Zhang$^{1}$,
    Zhuonan~Liu$^{1}$,
    Tomohito~Kawabata$^{1}$,
    Daeun~Song$^{2}$,
    Xuesu~Xiao$^{2}$,
    Ling Xiao$^{1,\dagger}$~\IEEEmembership{Senior Member,~IEEE}%
\thanks{$^\dagger$Corresponding author: \texttt{ling@ist.hokudai.ac.jp}.}%
\thanks{This work was supported by JSPS KAKENHI (Grant No. 24K20787).}%
\thanks{$^1$Graduate School of Information Science and Technology, Hokkaido University, Sapporo, Japan.}%
\thanks{$^2$Graduate School of Computer Science, George Mason University, Fairfax County, Virginia, USA}%
}
}

\maketitle

\begin{abstract}
Socially compliant navigation requires robots to move safely and appropriately in human-centered environments by respecting social norms. However, social norms are often ambiguous, and in a single scenario, multiple actions may be equally acceptable. Most existing methods simplify this problem by assuming a single ``correct'' action, which limits their ability to handle real-world social uncertainty.
In this work, we propose MAction-SocialNav, an efficient vision language model for socially compliant navigation that explicitly addresses action ambiguity, enabling generating multiple plausible actions within one scenario. To enhance the model’s reasoning capability, we introduce a novel meta-cognitive prompt (MCP) method. Furthermore, to evaluate the proposed method, we curate a multi-action socially compliant navigation dataset that accounts for diverse conditions, including crowd density, indoor and outdoor environments, and dual human annotations. The dataset contains 789 samples, each with three-turn conversation, split into 710 training samples and 79 test samples through random selection. We also design five evaluation metrics to assess high-level decision precision, safety, and diversity.
Extensive experiments demonstrate that the proposed MAction-SocialNav achieves strong social reasoning performance while maintaining high efficiency, highlighting its potential for real-world human robot navigation. 
Compared with zero-shot GPT-4o and Claude, our model achieves substantially higher decision quality (APG: 0.595 vs. 0.000/0.025) and safety alignment (ER: 0.264 vs. 0.642/0.668), while maintaining real-time efficiency (1.524 FPS, over 3× faster).
\end{abstract}


\section{Introduction}
\begin{figure}[t]
    \centering
    \includegraphics[width=\columnwidth]{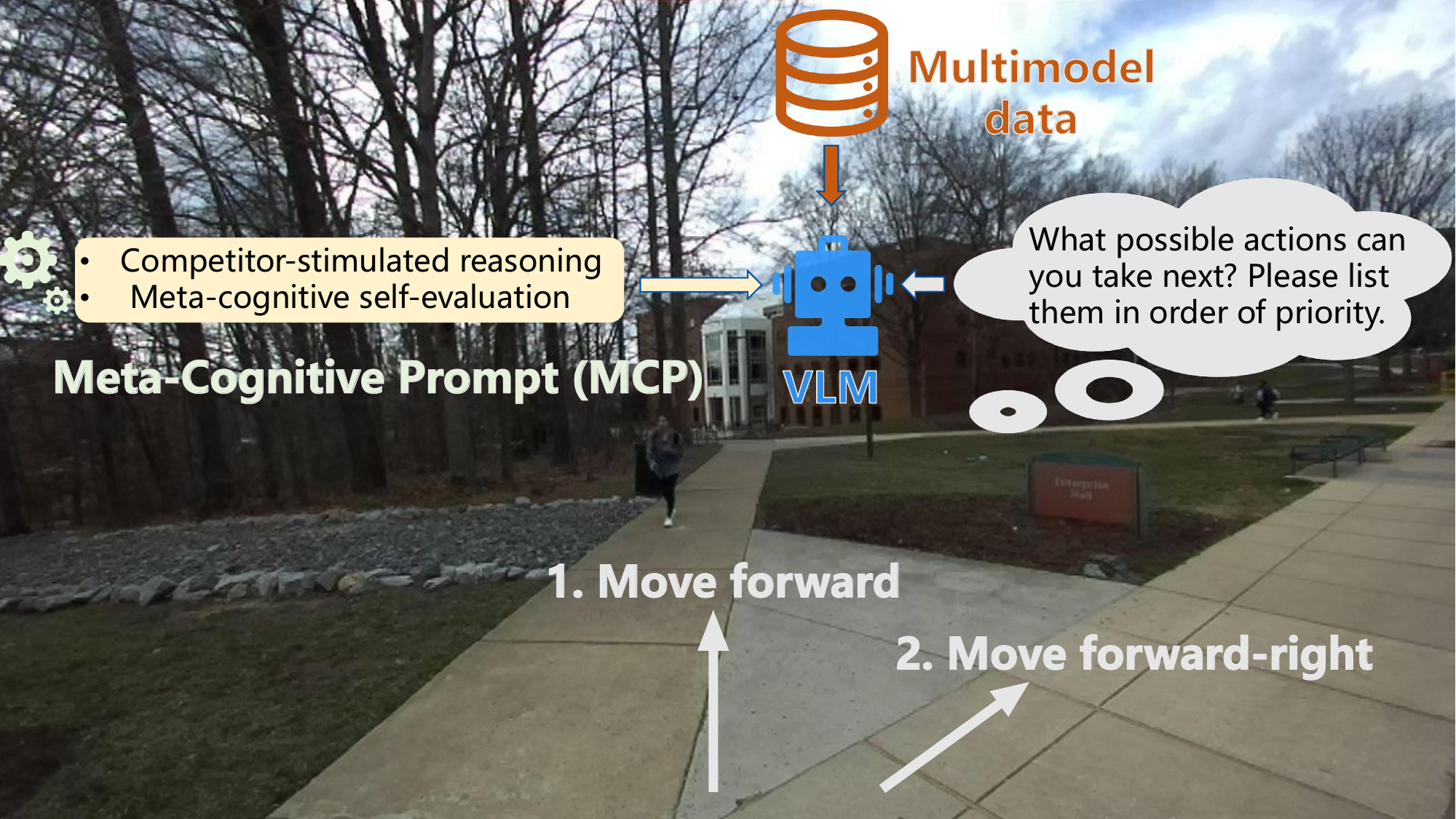}
    \caption{\textbf{Task formulation and high-level concept of multi-action social navigation.}
Given multimodal observations, the agent is required to reason about the scene and generate multiple feasible navigation actions ranked by priority, rather than a single deterministic action. A meta-cognitive prompt (MCP) is proposed to stimulate structured reasoning and self-evaluation in the vision language model (VLM), enabling socially compliant and interpretable decision-making.}
    \label{fig:overview}
\end{figure}

\IEEEPARstart{T}{he} proliferation of mobile robots in unstructured, human centric environments, ranging from service robots in retail hubs to delivery platforms on urban sidewalks, has elevated the importance of socially compliant navigation~\cite{raj2024rethinking}. While conventional industrial navigation focuses primarily on path efficiency and obstacle avoidance, robots operating in shared spaces with human must exhibit social compliance. This requires navigating beyond mere geometric collision avoidance to adhere to social norms, such as respecting personal space, yielding right-of-way, aligning with human traffic flows, and avoiding walking on grassed areas~\cite{singamaneni2024survey,mirsky2024conflict}. Therefore, the fundamental challenge in social navigation is to formulate a decision-making framework that prioritizes socially compliant actions over purely feasible ones.

Existing approaches~\cite{song2024vlm,sathyamoorthy2024convoi} predominantly formulate social navigation as a deterministic decision-making problem. However, in real-world social environments, pedestrians often exhibit multiple equally compliant behaviors within the same scene; for example, both stopping and bypassing an obstacle may be feasible choices in certain situations. By enforcing a single ground truth action, existing methods~\cite{kawabata2025socialnav,zu2024language} overlook this inherent diversity in human decision-making and penalize valid alternative behaviors. Such deterministic supervision restricts the model to a narrow solution space, preventing it from capturing the full distribution of socially acceptable interactions.

Furthermore, most existing approaches~\cite{weerakoon2025behav, xu2024mobility} rely on large scale vision language models (VLMs), which suffer from high inference latency, significantly restricting their deployment in real-world robotic systems.
Small language models (SLMs) have recently emerged as resource-efficient alternatives to computationally expensive large language models (LLMs) in various tasks. However, their potential has not yet been explored for socially compliant navigation, where efficiency, responsiveness, and reliable social reasoning are all critical.

To address the above-mentioned challenges, firstly, we propose MAction-SocialNav, a socially compliant navigation model capable of predicting multiple plausible actions in a given social environment. Then, to enhance social reasoning performance, we introduce a novel meta-cognitive prompt (MCP) method. Moreover, we construct a multi-action socially compliant navigation dataset and design five evaluation metrics to assess high-level decision precision, safety, and diversity of the proposed model.

The main contributions of this work are summarized as follows:
\begin{itemize}
    \item We propose a socially compliant navigation model that can predict multiple socially acceptable actions in a given environment. To the best of our knowledge, this is the first work to explicitly model and evaluate action ambiguity in socially compliant navigation.
    \item We introduce a novel MCP method, which comprises two components: meta-cognitive self-evaluation and competitor-stimulated reasoning. We also construct a multi-action dataset for evaluating the performance of proposed model. Moreover, We design five evaluation metrics to assess high-level decision precision, safety, and diversity.
    \item Comprehensive experiments demonstrate that our method significantly outperforms typical zero-shot large scale VLMs in generating safe and socially appropriate navigation actions in complex human–robot interaction scenarios.
\end{itemize}

\section{RELATED WORK}

\subsection{Social Robot Navigation}
Research on social robot navigation spans from classical hand-engineered models to modern learning-based frameworks. Traditional approaches, such as the social force model~\cite{helbing1995social} and proxemics-based methods~\cite{takayama2009influences,mumm2011human}, explicitly encode human motion patterns and interpersonal spatial constraints. While these methods are interpretable, their limited adaptability constrains performance in dynamic and diverse social environments.

To address these limitations, recent studies increasingly adopt data-driven approaches, including imitation learning for socially compliant trajectory generation~\cite{tai2018socially,ling2024socialgail}, deep reinforcement learning for optimizing long-horizon interactions~\cite{chen2017socially,chen2019crowd,kathuria2025learning}, and trajectory-prediction models for inferring human intent~\cite{salzmann2023robots,alahi2016social}. Graph neural networks (GNNs) further enhance multi-agent reasoning by explicitly modeling relational dependencies among humans and robots~\cite{bachiller2022graph}. Despite these advances, most learning-based methods focus primarily on generating feasible behaviors and often struggle to capture latent social norms, group dynamics, and nuanced human–robot relationships. As a result, achieving truly socially aware navigation remains an open challenge.

In parallel, recent progress in vision language navigation (VLN) demonstrates that language can serve as an effective interface for embodied navigation, enabling agents to ground high-level intent in perception and to generalize beyond hand-crafted cost functions~\cite{wu2024vision,chen2025constraint}. Building on this paradigm, several recent works extend VLMs to social navigation settings~\cite{song2024vlm, song2025vl, narasimhan2025olivia}. Notably, Social-LLaVA~\cite{payandeh2024social} introduces the SNEI dataset and formulates social navigation as a multi-stage process involving perception, prediction, chain-of-thought reasoning, action selection, and explanation. Models trained on SNEI, such as LLaVA-v1.5-7B, demonstrate promising capabilities in understanding social cues, anticipating pedestrian behavior, and generating interpretable navigation decisions.

However, three key limitations remain. First, real-world social navigation often admits multiple equally compliant actions rather than a single correct choice, requiring context-dependent evaluation instead of discrete classification. In contrast, SNEI provides only one annotated action per scenario, limiting its ability to model inherent social ambiguity. Second, the limited scale of SNEI (325 multi-turn dialogues) restricts the diversity of social situations and reduces the robustness and generalization of learned models. Third, existing work largely overlooks the efficiency of VLMs for socially compliant navigation. Although the task involves high-level reasoning, low inference latency remains critical for real-world robotic deployment.

\subsection{VLM Reasoning.}
LLMs have emerged as powerful tools across a wide range of domains due to their strong capabilities in understanding and generating human-like text. However, since vanilla LLMs are primarily trained for generic natural language processing objectives, they often perform poorly on tasks that require structured reasoning or domain-specific decision-making~\cite{rae2021scaling}. 

To address this limitation, Chain-of-Thought (CoT) prompting has been proposed to explicitly elicit intermediate reasoning steps, significantly improving the reasoning performance of LLMs by encouraging step-by-step deliberation. Building upon the linear reasoning paradigm of CoT, Tree-of-Thoughts (ToT)~\cite{yao2023tree} introduces a branching reasoning structure, where LLMs are instructed to simulate discussions among multiple experts and explore diverse reasoning paths before reaching a consensus. These prompting strategies substantially enhance reasoning accuracy, but they also incur considerable inference latency and computational overhead, which limits their applicability in real-time or resource-constrained scenarios.

Motivated by these challenges, SLMs have gained increasing attention as a promising alternative. SLMs are generally defined as models with substantially fewer parameters than large scale LLMs, typically not exceeding 3 billion parameters. Owing to their compact size, SLMs can be deployed on end-user devices such as personal computers and smartphones, even without GPU acceleration. Representative examples include the Phi series~\cite{gunasekar2023textbooks}, TinyLLaMA~\cite{zhou2024tinyllava}, and NVILA~\cite{liu2025nvila}. However, their potential for socially compliant robot navigation remains largely underexplored.

\section{Method}
\begin{figure*}[t!]
    \centering
    \includegraphics[width=0.95\linewidth]{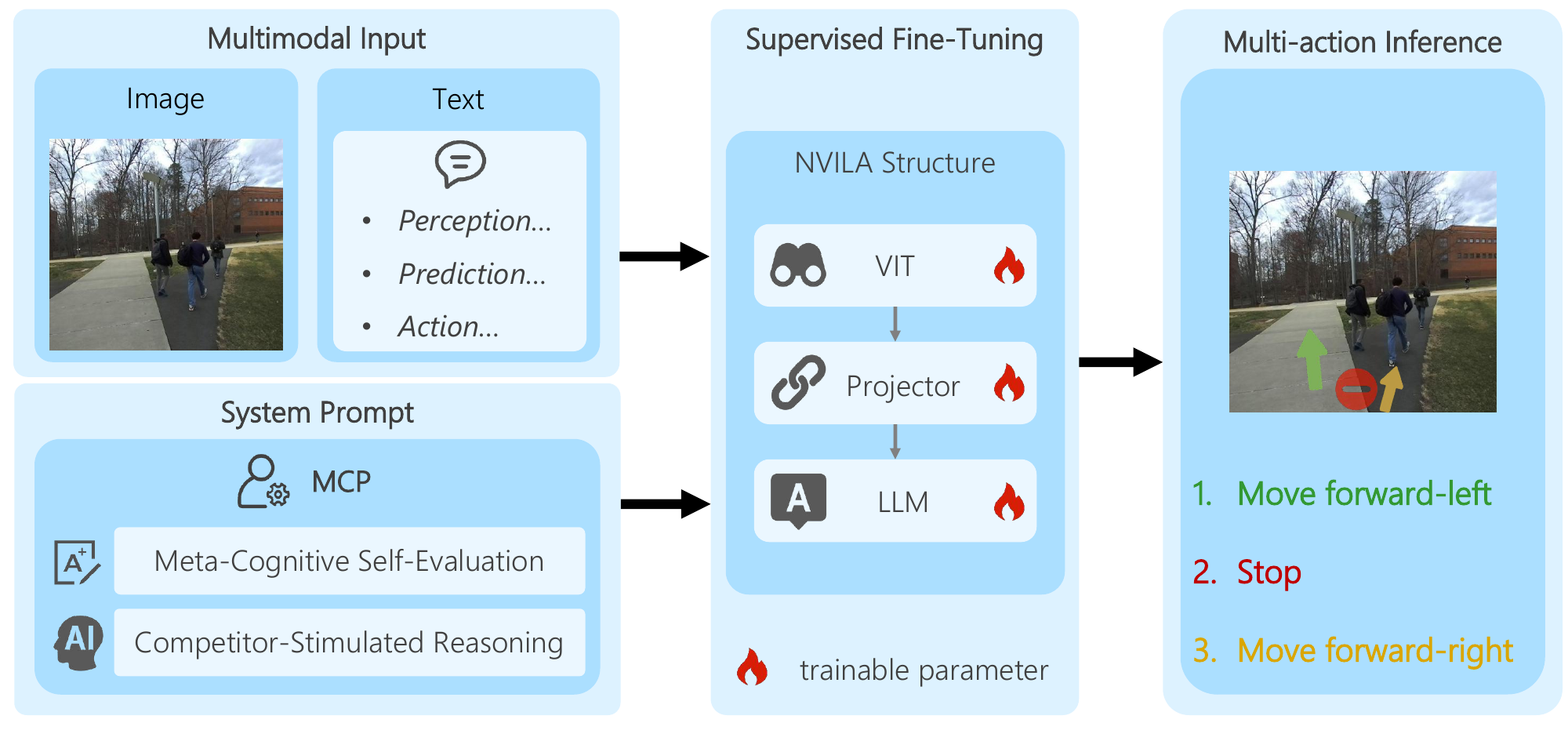}
    \caption{\textbf{Overview of MAction-SocialNav Framework.} We formulate socially compliant navigation as a multi-turn dialogue process. Given a scene observation $I$ and a designed MCP ($S_{\mathrm{MCP}}$) as the system prompt, the model sequentially performs perception and prediction through intermediate queries, and finally generates a ranked set of executable actions.}
    \label{fig:method_pipeline}
\end{figure*}

\subsection{Overview}
This paper proposes an efficient and effective VLM for multi-action socially compliant navigation. Fig.~\ref{fig:method_pipeline} illustrates the overall pipeline of the proposed MAction-SocialNav framework. We adopt the NVILA~\cite{liu2025nvila} architecture as the backbone and fine-tune it for multi-action social navigation. Specifically, we first reformulate supervision as multi-action ranking over a grounded discrete action space $\mathcal{A}$, enabling the model to learn preferences among plausible actions instead of imitating a single action. Then, we propose a novel meta-cognitive prompt (MCP) method to enhance reasoning capability under SLMs constraints. Finally, we curate a multi-action socially compliant navigation dataset and design five evaluation metrics to assess high-level decision precision, safety, and diversity. 

Given a visual observation $I$ and a multi-turn conversation $T$, the proposed meta-cognitive prompt (MCP) is used as the system prompt and concatenated with the multi-turn conversation $T$. The goal is to generate a textual response $Y$ that specifies the robot’s next behavior.

Formally, the model is trained via full parameter supervised fine-tuning (SFT). The optimization objective is the response sequence $Y=\{y_1,y_2,\dots,y_L\}$ ($L$ is the response length) via the autoregressive next-token prediction loss, conditioned on the injected system prompt:
\begin{equation}
    \mathcal{L} = - \sum_{t=1}^{T} \log p(y_t \mid I, S_{MCP}, T_{<t}, y_{<t}).
\end{equation}

This formulation ensures that the learned policy $p$ intrinsically incorporates the high-level constraints defined in $S_{\mathrm{MCP}}$ without requiring runtime search or external verifiers. Instead of predicting a single deterministic action, we model decision-making as ranking a set of discrete actions, which better reflects the inherent ambiguity of real-world social navigation. By leveraging the autoregressive dependency on previously generated tokens $y_{<t}$, the model implicitly enforces a hierarchical structure over actions. Conditioning each prediction on prior outputs allows the model to establish a preference ordering, producing a ranked set of socially compliant actions rather than an isolated decision. Details of the discrete action space and ranking protocol are introduced in Subsection~\ref{subsec:dataset}.

\begin{figure}[t]
    \centering
    \includegraphics[width=\linewidth]{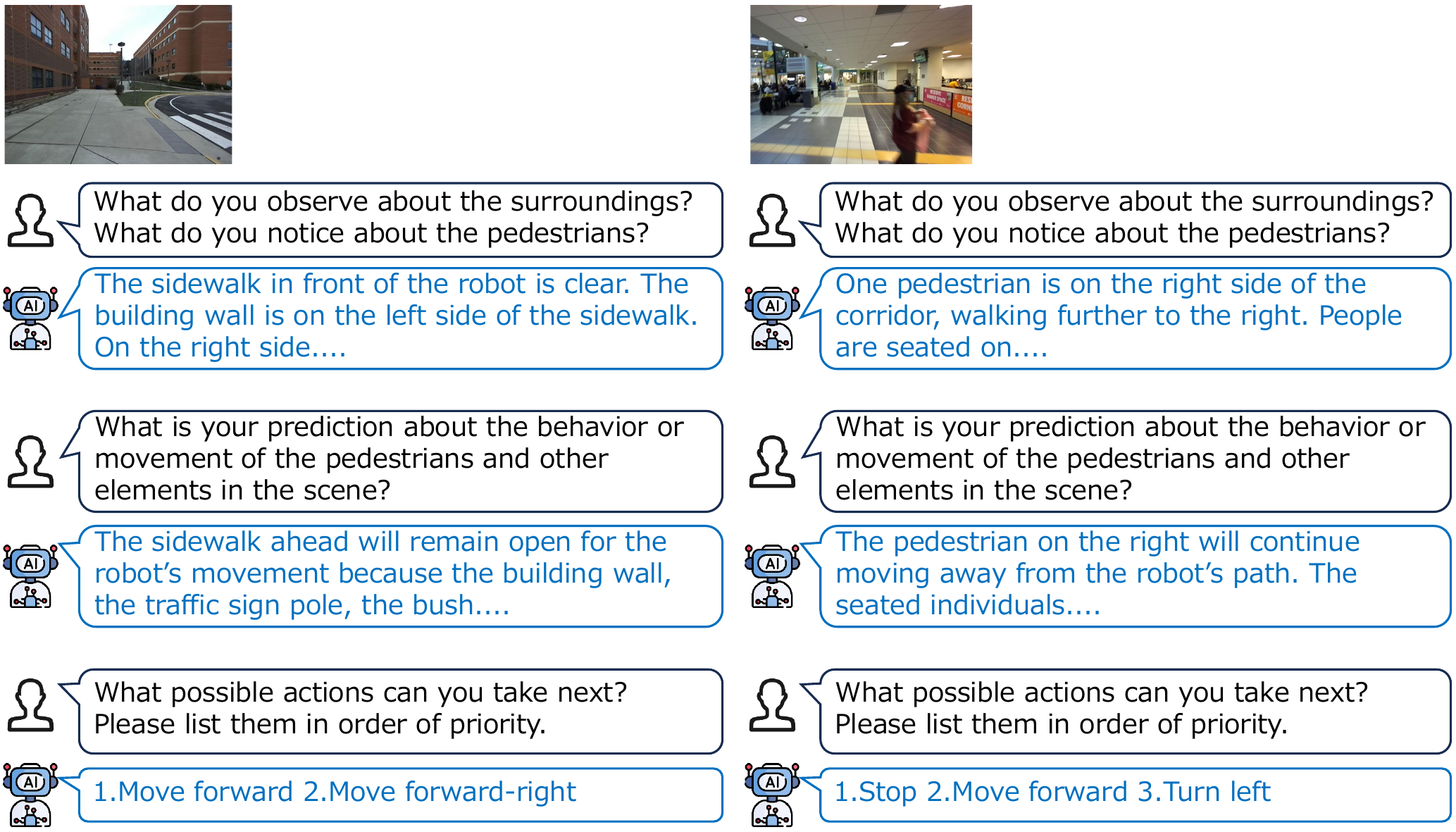}
    \caption{Multi-turn conversation dataset with action ranking. Each training sample consists of a visual observation paired with a multi-turn dialogue. All assistant responses are annotated by two human annotators. To support multi-action supervision, we introduce a hierarchical action ranking protocol. Specifically, we generate a ranked set of candidate actions based on a priority hierarchy: Feasibility$\rightarrow$ Social norms$\rightarrow$ Efficiency.}
    \label{fig:dataset_format}
\end{figure}

\subsection{Meta-Cognitive Prompt (MCP)}
\label{subsec:MCP}
We propose a novel meta-cognitive prompt method which consists of a \textbf{meta-cognitive self-evaluation} and \textbf{competitor-stimulated reasoning} (as shown in Table~\ref{tab:prompt_design}). 

\noindent\textbf{a) Meta-cognitive self-evaluation.}
We introduce a \emph{meta-cognitive self-evaluation} prompt $S_{\mathrm{meta}}$ that enforces an implicit recursive optimization loop; details are provided in Table~\ref{tab:prompt_design}. By explicitly introducing a \emph{scoring signal} and a \emph{decision threshold}, the model is encouraged to internally reallocate attention toward safety-critical and socially relevant factors (e.g., pedestrian proximity and trajectory conflicts) before committing to the final action decision.

\begin{table*}[t]
\centering
\caption{Details of the proposed MCP method, which is a combination of meta-cognitive self-evaluation and competitor-stimulated reasoning.}
\label{tab:prompt_design}
\begin{tabular}{p{4cm} p{12cm}}
\toprule
\textbf{Prompt Type} & \textbf{Prompt Description} \\
\midrule
\multirow{7}{*}{Meta-cognitive self-evaluation} &
Implement a silent, recursive self-evaluation loop. Before answering, internally generate a draft and score it based on strict safety and social adherence standards. Set 90 as the minimum passing threshold, but do NOT cap the score at 100. If a solution is exceptionally robust or you are highly confident, you are encouraged to assign a score exceeding 100. If the score is below 90, you must critically analyze the flaws, refine your logic, and simulate the outcome again. Repeat this internal iteration until the solution meets or exceeds the 90-point threshold. Output ONLY the final, optimized response without revealing the intermediate thinking steps.\\
\midrule
\multirow{3}{*}{Competitor-stimulated reasoning} &
You are an intelligent assistant specializing in socially compliant robot navigation.
You must understand human behaviors, infer intentions, and plan safe, smooth, and socially appropriate paths. You should perform competitively against \{humans, other AI models, other AI models like you.\}. \\
\bottomrule
\end{tabular}
\end{table*}

\textbf{b) Competitor-stimulated reasoning.}
Prior studies on LLMs suggest that defining a \emph{competitor} and reference standard can influence a model’s effort allocation and reasoning depth~\cite{zhou2022large}.
Motivated by this observation, we design a competitor-based system prompt $S_{\mathrm{com}}$ to examine their effects in the context of socially compliant robotic navigation. Specifically, $S_{\mathrm{com}}$ consists of three modes:
\begin{itemize}
    \item \textbf{Competing against humans ($S_{\mathrm{com}}^{\text{human}}$):}
    This setting encourages human-like intuition and socially grounded navigation behaviors.
    
    \item \textbf{Competing against other AI models ($S_{\mathrm{com}}^{\text{AI}}$):}
    This establishes a reference based on general intelligence and commonly observed AI-level performance.
    
    \item \textbf{Competing against self ($S_{\mathrm{com}}^{\text{self}}$):}
    This self-referential formulation implicitly encourages deeper internal search and self-improvement.
\end{itemize}

The final prompt $S_{\mathrm{MCP}}$ is constructed by concatenating $S_{\mathrm{meta}}$ and $S_{\mathrm{com}}$:
\begin{equation}
S_{\mathrm{MCP}} = \left\{ S_{\mathrm{meta}} \,\Vert\, S_{\mathrm{com}} \right\}.
\end{equation}

\subsection{Curated Multi-Action Dataset}
\label{subsec:dataset}
Existing social navigation datasets typically annotate each scene with a \emph{single} ground truth action accompanied by an explanation, which is not suitable for evaluation in our multi-action setting.
To address this limitation, we manually re-annotate each sample to support multi-action supervision with explicit ranking information. Moreover, unlike prior works that rely on open-ended textual descriptions, we define a grounded \textbf{discrete action space} $\mathcal{A}$ that consists of six motion primitives: 
\{\textit{Move forward}, \textit{Move forward-left}, \textit{Move forward-right}, \textit{Turn left}, \textit{Turn right}, and \textit{Stop}\}.
For each scene, we annotate a ranked subset of $\mathcal{A}$ using a strict hierarchical protocol, defined as follows:
\begin{enumerate}
    \item \textbf{Feasibility.} Actions that would cause collisions (e.g., with pedestrians or obstacles) or traverse non-drivable regions are removed.
    \item \textbf{Social norms.} Among feasible actions, those that maintain comfortable interpersonal distances are ranked higher.
    \item \textbf{Efficiency.} Among socially safe actions, those that maximize forward progress are preferred. Specifically, \textit{Move forward} is ranked above \textit{Move forward-left/right}, which are ranked above turning actions. \textit{Stop} is included only when necessary or as a fallback.
\end{enumerate}
This hierarchical annotation strategy shifts the learning objective from imitating a single action to modeling a preference distribution over plausible and socially appropriate behaviors.
Finally, we curate a dataset consisting of \textbf{789 samples}, with \textbf{79 samples} held out for testing. Each sample follows a three-stage interaction protocol:
(1) \textit{Scene Observation}, (2) \textit{Motion Prediction}, (3) \textit{Ranked Action Generation} as show in Fig.~\ref{fig:dataset_format}.
Importantly, we maintain the same input–output format and inference procedure during both training and inference, ensuring a consistent training–inference pipeline.

%\subsection{Training Objective}
%The model is trained via full parameter supervised fine-tuning (SFT). The optimization objective is the autoregressive next-token prediction loss, conditioned on the injected system prompt:
%\begin{equation}\mathcal{L} = - \sum_{t=1}^{T} \log p(y_t \mid I, S_{MCP}, T_{<t}, y_{<t}).
%\end{equation}
%This formulation ensures that the learned policy $p$ intrinsically incorporates the high-level constraints defined in $S_{\mathrm{MCP}}$ without requiring runtime search or external verifiers.

\subsection{Proposed Evaluation Metrics}
We design five evaluation metrics to assess high-level decision precision, safety, and diversity. 

\noindent\textbf{Pred@1} evaluates whether the top-ranked predicted action belongs to the ground truth (GT) acceptable action set and is defined as
\begin{equation}
\mathrm{Pred}@1 = \mathbb{I}[\hat{a}_1 \in \mathbf{a}].
\end{equation}
where $\hat{a}_1$ denotes the first action in the predicted action set, $\mathbf{a}$ represents the GT action set. 

\noindent\textbf{Pred@n} measures prediction precision by comparing each predicted action against the ground truth action set and is defined as
\begin{equation}
\mathrm{Pred}@n =
\frac{1}{|\hat{\mathbf{a}}|}
\sum_{i=1}^{|\hat{\mathbf{a}}|}
\left(
\mathbb{I}[\hat{a}_i \in \mathbf{a}]
-
\mathbb{I}[\hat{a}_i \notin \mathbf{a}]
\right).
\end{equation}
where $|\hat{\mathbf{a}}|$ donates the number of action of predicted action set.
This metric ranges from $-1$ to $1$, where higher values indicate cleaner predictions with fewer extraneous actions.

\noindent\textbf{All-Pred-in-GT (APG)} evaluates whether all predicted actions are valid according to the ground truth set and is defined as
\begin{equation}
\mathrm{APG} =
\mathbb{I}
\left[
\forall \hat{a}_i \in \hat{\mathbf{a}},\;
\hat{a}_i \in \mathbf{a}
\right].
\end{equation}
This metric enforces a strict precision constraint and equals $1$ only when the predicted action set is a subset of the ground truth.

\noindent\textbf{Multi-action accuracy (MAA)} captures ordering sensitivity in multi-step predictions by assigning higher importance to earlier actions. Let $w = [6,5,4,3,2,1]$ denote a predefined weight vector. The metric is defined as
\begin{equation}
\mathrm{MAA} =
\frac{1}{|\mathbf{a}|}
\sum_{i=1}^{\min(|\hat{\mathbf{a}}|,6)}
w_i \cdot \mathbb{I}[\hat{a}_i \in \mathbf{a}],
\end{equation}
and is set to $0$ when $|\hat{\mathbf{a}}| > |\mathbf{a}|$. 
where $|\mathbf{a}|$ donates the number of action of predicted action set.
This metric emphasizes early correct decisions, which are critical in sequential decision-making tasks.

\noindent\textbf{Error rate (ER)} measures the proportion of hallucinated actions among all predicted actions and is defined as
\begin{equation}
\mathrm{ER} =
\frac{1}{|\hat{\mathbf{a}}|}
\sum_{i=1}^{|\hat{\mathbf{a}}|}
\mathbb{I}[\hat{a}_i \notin \mathbf{a}].
\end{equation}
This metric ranges from $0$ to $1$, where higher values indicate a stronger tendency to generate invalid actions.

\section{Experiment}

\subsection{Experimental Setup}

% \textbf{Backbone and Training Configuration.} 
We implement our framework using the NVILA-Lite-2B~\cite{liu2025nvila} architecture. We fine-tune the vision projector and LoRA on 4 NVIDIA RTX 8000 GPUs (48GB memory each) with a learning rate of $1 \times 10^{-4}$. 
We set 2 as batch size of each GPU and apply gradient accumulation with 4 steps, resulting in an effective global batch size of 64. 
Training is conducted for 10 epochs using DeepSpeed ZeRO-2 and $BF16$ precision.

\subsection{Main Results}
\begin{table}[t]
\centering
\caption{Comparison of MAction-SocialNav with and without MCP, and with causal reasoning.
Experimental results show that the proposed MCP significantly outperforms both the variant without MCP and the variant equipped with causal reasoning, demonstrating the superior effectiveness of MCP for multi-action social navigation.}
\label{tab:main_results}
\resizebox{0.5\textwidth}{!}{
\begin{tabular}{l|ccccc}
\toprule
\textbf{Method} 
& {Pred@1}$\uparrow$ & {Pred@n}$\uparrow$ & {APG}$\uparrow$ & {MAA}$\uparrow$ & {ER}$\downarrow$ \\
\midrule
w/o MCP
& 0.734 & 0.389 & 0.532 & 3.048 & 0.306 \\
w/ Causal Reasoning
& 0.747 & 0.351 & 0.532 & 3.202 & 0.325 \\
w/ MCP
& \textbf{0.760} & \textbf{0.473} & \textbf{0.595} & \textbf{3.571} & \textbf{0.264} \\ 
\bottomrule
\end{tabular}
}
\end{table}

As shown in Table~\ref{tab:main_results}, introducing explicit reasoning does not consistently improve performance and, in several cases, even leads to degradation. Here, MAction-SocialNav (w/ Causal Reasoning) denotes a baseline pipeline that does not employ any system prompt, but instead inserts an additional reasoning turn before final action generation. This reasoning stage follows a causal reasoning paradigm, where the model explicitly derives a sequence of high-level socially compliant actions by interpreting perception and prediction cues through a human-like CoT process.

In contrast, MAction-SocialNav with MCP relies solely on the proposed MCP as the system prompt, without introducing any explicit intermediate reasoning steps. The results demonstrate that MCP consistently outperforms causal reasoning, indicating that meta-cognitive guidance via system prompting is more effective than explicit reasoning for socially compliant multi-action navigation.

\subsection{Comparison with Zero-Shot Large Multimodal Models}

To examine whether general-purpose large multimodal models can satisfy the strict executability, safety, and latency requirements of socially compliant navigation without task-specific alignment, we compare our method with two representative closed-source models, GPT-4o and Claude. Both models are evaluated in a zero-shot setting and are not fine-tuned on the navigation task.

To ensure a fair comparison, GPT-4o and Claude are queried using the same constrained prompt (see Fig.~\ref{fig:constrained_prompt}), which explicitly encodes the action ranking protocol adopted in our framework, including feasibility pruning, safety prioritization, and efficiency-based ordering. No CoT prompting, additional reasoning steps, or post-processing are applied.

\begin{figure}[t]
\centering
\fbox{%
\begin{minipage}{0.95\linewidth}
\small
\textbf{Constrained prompt used in tuning GPT-4o and Claude.}

\vspace{4pt}
Given the current observation scenario, as a social robot, first prune infeasible actions, then select all executable actions from the following six actions:
\textit{Move forward, Move forward-left, Move forward-right, Turn left, Turn right, Stop}.

Rank the selected actions in descending priority according to:
(1) Social Safety, (2) Efficiency.

You may output between 1 and 6 actions depending on feasibility.
Output exactly one line using the following format, without any explanation or extra text:

\texttt{1.<action> 2.<action> ...}
\end{minipage}}
\caption{Constrained prompt used for querying GPT-4o and Claude.}
\label{fig:constrained_prompt}
\end{figure}

% \begin{quote}
% \small
% \textit{"Given the current observation scenario, as a social robot, first prune infeasible actions, then select all executable actions from the following 6 actions (Move forward, Move forward left, Move forward right, Turn left, Turn right, Stop). Rank the selected actions in descending priority according to: (1) Social Safety, (2) Efficiency. You may output between 1 and 6 actions depending on feasibility. Output exactly one line using the following format, without any explanation or extra text: '1.$<action>$ 2.$<action>$ ...'."}
% \end{quote}

\begin{table}[t]
\centering\
\caption{Comparison with Zero-Shot Large Multimodal Models.}
\label{tab:llm_comparison}
\resizebox{0.5\textwidth}{!}{
\begin{tabular}{l|cccccc}
\toprule
Method & Pred@1$\uparrow$ & Pred@n$\uparrow$ & APG$\uparrow$ & MAA$\uparrow$ & ER$\downarrow$ & FPS$\uparrow$ \\
\midrule
Claude (Zero-shot) & 0.570 & -0.349  & 0.025 & 0.230 & 0.668 & 0.452 \\
GPT-4o (Zero-shot) & 0.405 & -0.424  & 0.000 & 0.059 & 0.642 & 0.314 \\
\midrule
MAction-SocialNav & \textbf{0.760} & \textbf{0.473} & \textbf{0.595} & \textbf{3.571} & \textbf{0.264} & \textbf{1.524} \\
\bottomrule
\end{tabular}
}
\end{table}

\begin{figure*}[t!]
    \centering
    \includegraphics[width=0.85\linewidth]{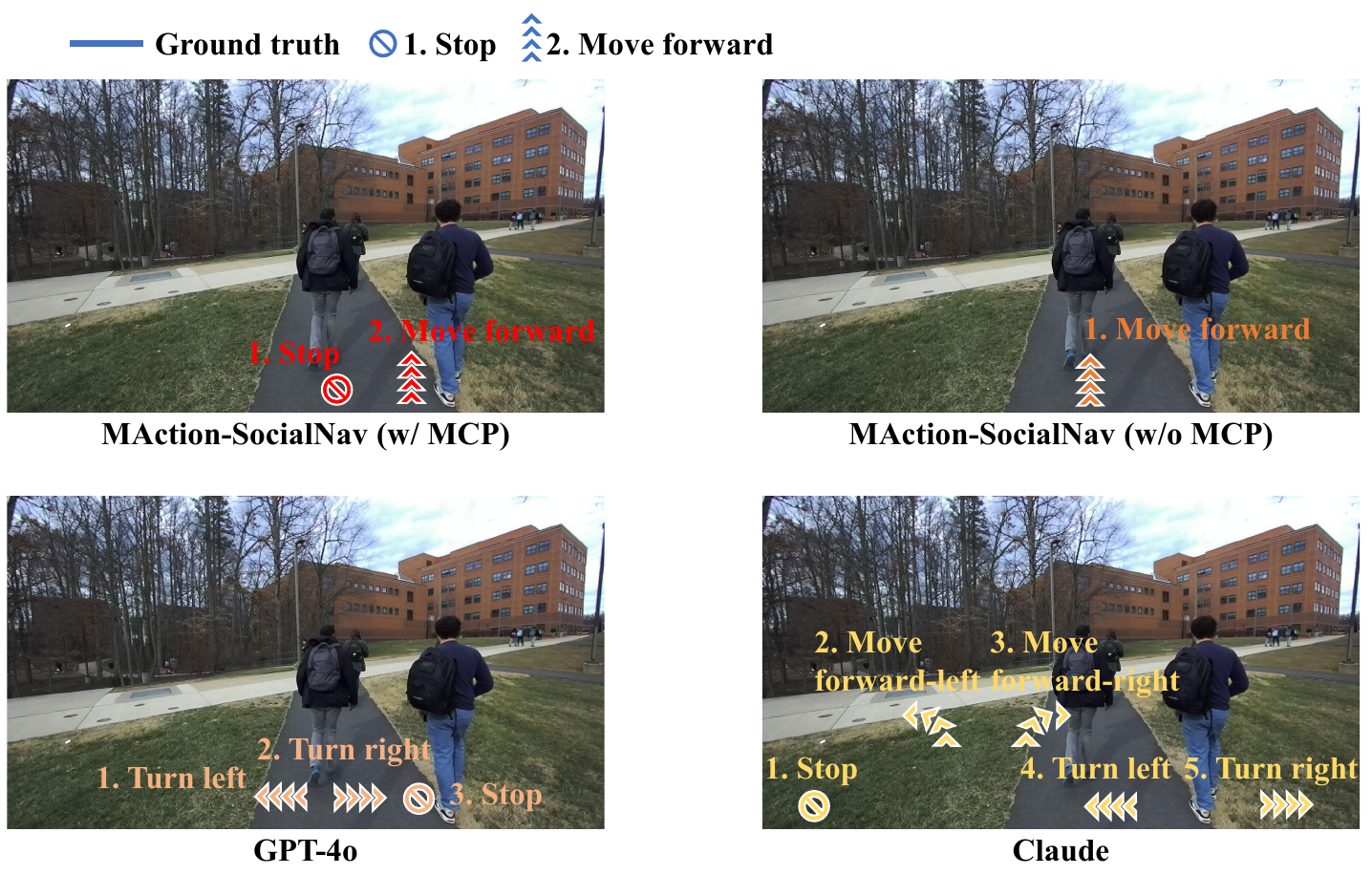}
    \caption{Visual comparison with GPT-4o, Claude, MAction-SocialNav without MCP, and MAction-SocialNav with MCP.
MAction-SocialNav with MCP predicts exactly the same set of feasible actions as the ground truth.}
    \label{fig:visual}
\end{figure*}

Table~\ref{tab:llm_comparison} summarizes the quantitative results. Fig.~\ref{fig:visual} shows the visual comparison with GPT-4o, Claude, MAction-SocialNav without MCP, and MAction-SocialNav with MCP. Overall, both GPT-4o and Claude perform poorly on this task, even when explicitly constrained by detailed prompts that encode feasibility and safety norms.

In terms of decision quality, zero-shot large VLMs exhibit limited performance in precision, safety, and diversity. As shown in Table~\ref{tab:llm_comparison}, Claude achieves a Pred@1 score of 0.570, while GPT-4o performs even worse with a Pred@1 of 0.405, indicating weak top-ranked action precision under the multi-action setting. In addition, both models show negative Pred@n scores, suggesting poor coverage of feasible actions. Their low MAA values further reflect limited diversity and insufficient multi-action awareness, while relatively high ER values indicate a higher risk of unsafe or socially inappropriate decisions. In contrast, MAction-SocialNav significantly outperforms both zero-shot baselines across all metrics. It achieves the highest Pred@1 (0.760), Pred@n (0.473), and APG (0.595), demonstrating superior precision and coverage in multi-action decision-making. Moreover, MAction-SocialNav substantially improves diversity-related metrics, with an MAA of 3.571, while reducing the error rate to 0.264. Notably, it also achieves the highest inference speed (1.524 FPS), highlighting its effectiveness and efficiency for socially compliant navigation.

\subsection{Ablational Studies}

\begin{table*}[t]
\centering
\caption{Ablational studies. We analyze the contributions of the proposed MCP, including meta-cognitive self-evaluation and competitor-stimulated reasoning with different competitors (humans, other AI models, and itself).
The results show that incorporating both meta-cognitive self-evaluation and competitor-stimulated reasoning can consistently improves performance. Combining meta-cognitive self-evaluation and competitor-stimulated reasoning with ``competing against other AI models'' ($S^{AI}_{com}$) achieves the best overall results.}
\label{tab:ablation_full}
\resizebox{0.85\textwidth}{!}
{
\begin{tabular}{cccc|ccccc}
\toprule
\multicolumn{4}{c|}{\textbf{MCP}}
& \multirow{3}{*}{Pred@1$\uparrow$}
& \multirow{3}{*}{Pred@n$\uparrow$}
& \multirow{3}{*}{APG$\uparrow$}
& \multirow{3}{*}{MAA$\uparrow$}
& \multirow{3}{*}{ER$\downarrow$} \\
\multirow{2}{*}{\makecell{Meta-cognitive\\self-evaluation}}
& \multicolumn{3}{c|}{Competitor-stimulated reasoning}\\
& $S_{\mathrm{com}}^{\text {human}}$
& $S_{\mathrm{com}}^{\text {self}}$
& $S_{\mathrm{com}}^{\text {AI}}$ \\

\midrule
-- & -- & -- & --
& 0.734 & 0.389 & 0.532 & 3.048 & 0.306 \\

\checkmark & -- & -- & --
& \textbf{0.785} & \underline{0.395}  & \underline{0.544} & \underline{3.137} & \underline{0.303} \\

-- & \checkmark & -- & --
& \textbf{0.785} & 0.356 & 0.468 & 2.772 & 0.322 \\

-- & -- & \checkmark & --
& \underline{0.760} & 0.367 & 0.468 & 3.052 & 0.317 \\

-- & -- & -- & \checkmark
& \underline{0.760} & 0.392 & 0.494 & 3.135 & 0.304 \\

\checkmark & \checkmark & -- & --
& 0.722 & 0.392 & 0.494 & 3.103 & 0.304 \\

\checkmark & -- & \checkmark & --
& 0.747 & 0.378 & 0.506 & 3.012 & 0.311 \\

\checkmark & -- & -- & \checkmark
& \underline{0.760} & \textbf{0.473} & \textbf{0.595} & \textbf{3.571} & \textbf{0.264} \\
\bottomrule
\end{tabular}
}
\end{table*}

To validate the efficacy of the proposed MCP, we conducted a comprehensive ablation study to analyze the individual and combined contributions of its core components: meta-cognitive self-evaluation and competitor-stimulated reasoning. As reported in Table~\ref{tab:ablation_full}, the baseline (Row 1) yields a Pred@1 of 0.734 and an MAA of 3.048. Incorporating Meta-cognitive self-evaluation alone (Row 2) results in a substantial gain in Pred@1 ($0.734 \rightarrow 0.785$), demonstrating the importance of the self-evaluation mechanism. Similarly, introducing competitor-stimulated reasoning in isolation (Rows 3-5) improves performance, with the $S_{\mathrm{com}}^\text{human}$ variant matching the peak Pred@1 score. However, single-module improvements often lack consistency across all metrics (e.g., Pred@n and APG).The results further reveal that the full potential of MCP is unlocked through the coupling of these modules. While combining self-evaluation with $S_{\mathrm{com}}^\text{human}$ or $S_{\mathrm{com}}^{\text{self}}$ yields moderate gains, the integration of Meta-cognitive self-evaluation with Competitor-stimulated reasoning against other AI models ($S_{\mathrm{com}}^\text{AI}$) achieves superior performance (Row 8). This configuration outperforms all other settings across comprehensive metrics, boosting Pred@n to 0.473 and MAA to 3.571, while significantly reducing the ER to 0.264. This empirical evidence confirms the synergistic effect of combining self-evaluation with competitive pressure from diverse AI counterparts, leading to more robust reasoning capabilities.

\subsection{Performance Across Scenario Difficulty Levels}
\begin{figure}[t!]
    \centering
    \includegraphics[width=\linewidth]{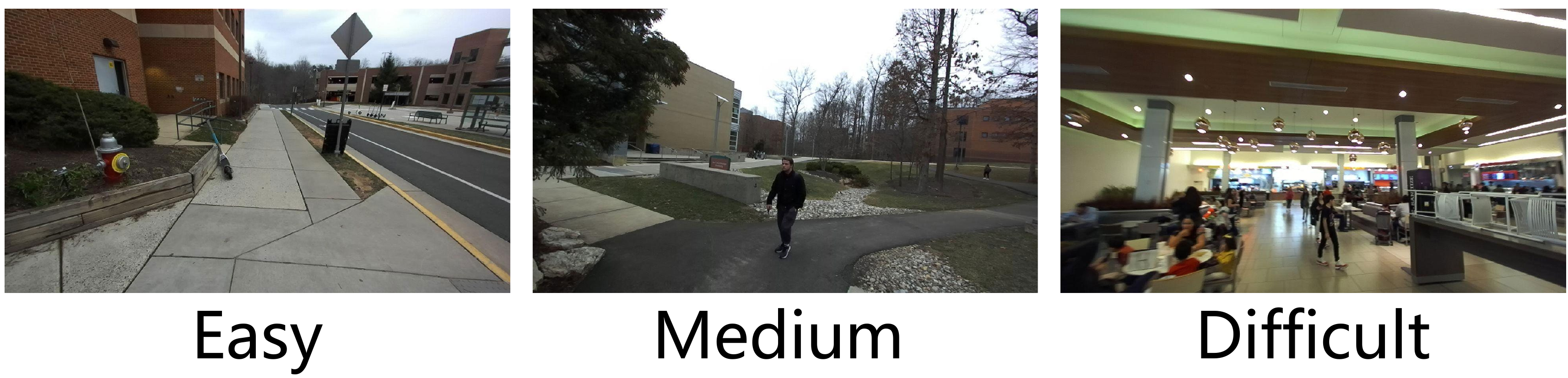}
    \caption{Examples of the test set across different difficulty levels.}
    \label{fig:easy-medium-difficult}
\end{figure}

To better understand how the proposed action ranking framework behaves under varying levels of scenario complexity, we further analyze model performance across different difficulty levels. The test set is partitioned into three subsets: \textit{Easy}, \textit{Medium}, and \textit{Difficult}, based on a predefined rule-based protocol (as shown in Fig.~\ref{fig:easy-medium-difficult}). The difficulty categorization considers three factors: (1) the complexity of the road network (e.g., whether multiple route options are available), (2) pedestrian complexity (e.g., whether a large number of pedestrians influence action selection), and (3) environmental complexity (e.g., cluttered surroundings or static obstacles).

Table~\ref{tab:difficulty_compare} summarizes the performance across the three difficulty levels. For models without the proposed MCP, evaluation metrics consistently degrade as scenario difficulty increases, reflecting the growing ambiguity and complexity in action selection. In particular, Medium and Difficult scenarios require finer-grained preference reasoning among multiple feasible actions, which unguided models struggle to handle reliably.
Conversely, the incorporation of MCP yields a distinct performance trajectory, diverging from the previously observed monotonic degradation. While performance in easy scenarios remains comparable, models leveraging MCP demonstrate superior stability in moderate and complex settings. Notably, in terms of $\text{Pred}@1$, performance in difficult scenarios actually surpasses that in medium ones. These results suggest that MCP instills robust behavioral priors via self-validation and competitive mechanisms, thereby not only relying on superficial visual correlations. and effectively optimizing the trade-off between safety and feasibility during complex human-robot interactions.

Overall, these results indicate that scenario difficulty amplifies the limitations of unguided decoding, whereas the proposed MCP alters the scaling behavior with respect to complexity, enabling more robust action ranking under challenging social navigation conditions.
\begin{table}[t]
\centering
\caption{Comparison of MAction-SocialNav across different scenario difficulty levels.
Experimental results show that the proposed MCP effectively enhances the performance of the model in complex social navigation environments, demonstrating the effectiveness of MCP in multi-action social navigation.}
\label{tab:difficulty_compare}
\resizebox{0.5\textwidth}{!}{
\begin{tabular}{l|c|ccccc}
\toprule
\textbf{Method}
& \textbf{Difficulty}
& Pred@1$\uparrow$  & Pred@n$\uparrow$ & APG$\uparrow$ & MAA$\uparrow$ & ER$\downarrow$ \\
\midrule
\multirow{3}{*}{\textbf{w/o MCP}} 
& Easy      
& 0.966 & 0.897 & 0.931 & 5.690 & 0.052 \\
& Medium    
& 0.621 & 0.193 & 0.414 & 1.724 & 0.403 \\
& Difficult 
& 0.571 & -0.041 & 0.143 & 1.226 & 0.521 \\
\midrule
\multirow{3}{*}{\textbf{w/ MCP}}
& Easy      
& 0.931 & 0.897 & 0.931 & 5.483 & 0.052 \\
& Medium    
& 0.621 & 0.276 & 0.483 & 2.702 & 0.362 \\
& Difficult 
& 0.714 & 0.159 & 0.286 & 2.131 & 0.421 \\
\bottomrule
\end{tabular}
}
\end{table}

\subsection{Limitations and Future Work}
Our current action space considers six discrete actions, which, while sufficient for stable and interpretable decision-making, may be insufficient for real-world deployment where finer-grained control (e.g., precise speed or trajectory adjustments) is required. In addition, our model focuses on high-level semantic reasoning and action selection, and is not designed to directly generate low-level control commands. As such, it should be integrated with a conventional local motion planner for real-world execution.

In future work, we plan to combine our high-level multi-action reasoning model with a low-level motion planner to enable full robot deployment. This hierarchical integration will allow us to evaluate safety-related metrics (e.g., collision rate, minimum distance to pedestrians) and efficiency metrics (e.g., path length and travel time) in realistic navigation scenarios.

\section{Conclusion}
In this paper, we propose MAction-SocialNav, a socially compliant navigation model that explicitly addresses action ambiguity in dynamic social environments. To enhance the reasoning capability of the proposed MAction-SocialNav, we introduce a novel meta-cognitive prompt (MCP). We further curate a dedicated multi-action social navigation dataset and design five evaluation metrics to assess high-level decision-making performance in terms of precision, safety, and diversity. Experimental results demonstrate that MAction-SocialNav substantially outperforms zero-shot large multimodal baselines, offering a practical and efficient pathway toward socially compliant robot navigation in complex real-world environments.

\bibliographystyle{IEEEtran}
\bibliography{refs}

\end{document}